\documentclass[runningheads]{llncs}

\usepackage{graphicx}
\usepackage{booktabs}
\usepackage{amsmath}
\usepackage{amssymb}
\usepackage{algorithm}
\usepackage{algpseudocode}
\usepackage{array}
\usepackage{tikz}
\usetikzlibrary{shapes.geometric, arrows.meta, positioning, calc}
\usepackage{marvosym}      
\usepackage[hidelinks]{hyperref}

\begin{document}

\title{ReviewSense: Cluster-Guided Evidence Selection for Actionable Review-to-Advice Generation}
\titlerunning{Cluster-Guided Actionable Advice from Customer Reviews}

\makeatletter
\newcommand{\printfnsymbol}[1]{%
  \textsuperscript{\@fnsymbol{#1}}%
}
\makeatother

\author{%
Siddhartha Krothapalli\inst{1}
\thanks{Equal contribution.}
\and
Kartikey Singh Bhandari\inst{2}
\printfnsymbol{1}
\and
Tridib Kumar Das\inst{3}
\and
Praveen Kumar\inst{4}
\and
Naveen Suravarpu\inst{4}
\and
Pratik Narang\inst{2}\textsuperscript{\Letter} 
}
\authorrunning{S. Krothapalli et al.}
\institute{%
Department of Mechanical Engineering, BITS Pilani, Pilani, India\\
\email{f20212207@pilani.bits-pilani.ac.in}
\and
Department of CSIS, BITS Pilani, Pilani, India\\
\email{\{p20241006,pratik.narang\}@pilani.bits-pilani.ac.in}
\and
Department of CSE, KIIT, Bhubaneswar, India\\
\email{2128106@kiit.ac.in}
\and
Birdeye Inc., Palo Alto, California, USA\\
\email{\{praveen.kumar1,Naveen.Suravarpu\}@birdeye.com}
}

\maketitle

\begin{abstract}
Customer reviews are a rich but weakly structured source of operational feedback. We study ReviewSense, a cluster-guided evidence-selection workflow that converts review corpora into action-oriented advice for human decision support. Reviews are embedded with a sentence transformer, grouped by iterative cosine-similarity clustering with threshold decay, and reduced to compact evidence before Large Language Model (LLM) generation. Evaluating on dental and healthcare service reviews with expert judgment of actionability and annotator-based theme alignment, we find that a single central representative preserves the dominant cluster theme in 77.3\% of cases but only 60.6\% of secondary themes: central evidence captures main issues more reliably than minority concerns. Against fixed-generator LLM baselines, five cluster representatives yield the advice closest to expert recommendations while using a fraction of the raw-review context, whereas a single representative is not the strongest front-end. Across three domains the clustering assigns every review while remaining competitive with standard algorithms. Our contribution is therefore an empirical account of evidence selection as the controlling design variable in review-to-advice generation, delivered as an inspectable human-in-the-loop workflow rather than a new clustering algorithm or a general review summarizer.
\end{abstract}

\section{Introduction}
\label{sec:intro}

Online reviews are a major source of market intelligence because they capture customer experiences at scale across products, services, and service encounters. Prior work shows that review text can reveal sentiment, salient attributes, and drivers of perceived quality~\cite{li2019effect,mustak2024using,pang-etal-2002-thumbs,wang2018topic}. In practice, however, many review-analysis pipelines still stop at sentiment labels, topic summaries, or descriptive dashboards~\cite{mustak2024using,praveen2024crafting,teichert2026reviews}. For managers, this creates a translation gap: large volumes of heterogeneous feedback are available, but the final step from recurring complaints to specific, implementable action is still largely manual.

This paper addresses that gap with \textit{ReviewSense}, a cluster-guided workflow for studying what evidence an LLM should see when turning review corpora into action-oriented advice. The framework first embeds reviews using Sentence-BERT~\cite{reimers-gurevych-2019-sentence}, then extracts recurrent complaint clusters using an iterative cosine-similarity procedure with threshold decay. It then selects compact evidence from the clusters, ranging from one central review to a small set of representatives, and passes that evidence to a LoRA-adapted Llama 3 8B~\cite{grattafiori2024llama,hu2022lora} model that generates issue-advice pairs. The design is intentionally structured: rather than prompting a language model directly on a large batch of raw reviews, we first isolate coherent issues so that generation is grounded in focused, inspectable evidence.

We evaluate ReviewSense on dental and healthcare review sets using two complementary forms of assessment. First, annotators assess whether generated outputs preserve dominant and secondary themes of the originating clusters. Second, expert reviewers judge whether the advice is concrete enough to support a plausible managerial action. The results show a useful boundary: central single-review conditioning captures dominant themes, but compact multi-representative evidence gives a stronger quality-context trade-off against LLM baselines. We therefore position ReviewSense as a human-centered decision-support workflow rather than as a new general-purpose clustering algorithm or a benchmark-oriented summarization system. We do not claim that review-to-action is a new problem, nor that any component is new: turning feedback into business intelligence is long-standing~\cite{lepenioti2020prescriptive,Pang2008OpinionMA}, and the encoder, clustering heuristic, and adapter are off-the-shelf. The contribution lies elsewhere: we treat \emph{evidence selection} as the controlling design variable in review-to-advice generation and measure it under a fixed generator, separating the front-end from generator capability. This distinguishes ReviewSense from retrieval-augmented generation (evidence is fixed by corpus structure, not retrieved per query), from opinion summarization (the target is a prescriptive issue-advice pair with a mechanism and an owner, not a consensus summary), and from actionable-insight extraction (which classifies feedback rather than generating and scoring advice). The account is deliberately two-sided: it reports where cluster-guided selection helps and where it fails.

The main contributions of this paper are as follows:
\begin{itemize}
    \item We formulate review-to-advice generation as a decision-support task in which weakly structured customer feedback is converted into issue-oriented managerial guidance.
    \item We present a practical pipeline that combines iterative semantic clustering, cluster-guided evidence selection, and LLM-based issue-advice generation, and we describe each component precisely enough to be reproduced (Algorithm~\ref{alg:cluster}).
    \item We compare evidence-selection front-ends under a fixed generator and show that compact multi-representative evidence gives the best quality-context trade-off, while single representatives remain useful as an interpretable ablation.
    \item We provide empirical observations from domain-specific datasets using expert-based evaluation of advice quality, and we analyze a concrete failure mode: secondary themes within multi-issue clusters are captured less reliably than dominant ones.
\end{itemize}

\begin{figure}[t]
    \centering
    \begin{tikzpicture}[
      node distance=4mm and 5mm,
      box/.style={draw, rounded corners, align=center, minimum height=8.5mm, text width=15mm, font=\scriptsize, fill=gray!8},
      io/.style={draw, align=center, minimum height=8.5mm, text width=14mm, font=\scriptsize, fill=blue!6},
      arr/.style={-{Latex[length=1.8mm]}, semithick}
    ]
    \node[io] (rev) {Review\\corpus};
    \node[box, right=of rev] (emb) {SBERT\\embedding};
    \node[box, right=of emb] (clu) {Iterative\\similarity\\clustering};
    \node[box, right=of clu] (rep) {Evidence\\selection};
    \node[box, right=of rep] (gen) {LoRA\\Llama-3\\generation};
    \node[io, right=of gen] (out) {Issue-advice\\pairs};
    \draw[arr] (rev)--(emb);
    \draw[arr] (emb)--(clu);
    \draw[arr] (clu)--(rep);
    \draw[arr] (rep)--(gen);
    \draw[arr] (gen)--(out);
    \end{tikzpicture}
    \caption{ReviewSense pipeline. Reviews are embedded, grouped into recurrent complaint clusters, reduced to compact representative evidence, and converted into issue-advice pairs for decision support.}
    \label{fig:pipeline}
\end{figure}

\section{Related Work}
\label{sec:related}

\subsection{Review Mining, Aspect Analysis, and Summarization}
Opinion mining moved from document-level polarity~\cite{pang-etal-2002-thumbs} to feature- and aspect-level extraction of opinion-bearing content~\cite{brody-elhadad-2010-unsupervised,1014073,popescu-etzioni-2005-extracting}, framed broadly as review search, summarization, and business intelligence~\cite{Pang2008OpinionMA}, yet its outputs stay descriptive. Aspect-based sentiment analysis adds aspect terms, categories, and polarity, with SemEval-2014 benchmarks~\cite{pontiki-etal-2014-semeval} and BERT sentence-pair formulations~\cite{devlin-etal-2019-bert,sun-etal-2019-utilizing}, but still returns labels rather than operational recommendations. The closest task, opinion summarization, condenses reviews into shared-opinion summaries using weakly supervised, unsupervised, customizable, aspect-controllable, long-form, or aspect-guided methods~\cite{amplayo-etal-2021-aspect,angelidis-lapata-2018-summarizing,bhaskar2022prompted,boytsov-etal-2025-end,brazinskas-etal-2020-unsupervised,nayeem2024lfosum,suhara2020opiniondigest}. ReviewSense instead targets issue-advice generation, judged by whether outputs propose specific, evidence-grounded, feasible actions.

\subsection{Structure Discovery and LLM-based Generation}
Structure-discovery baselines include LDA~\cite{blei2003latent}, review-oriented topic and sentiment-topic models~\cite{li2019effect,wang2018topic}, contextualized topic models~\cite{bianchi-etal-2021-pretraining}, SBERT clustering~\cite{reimers-gurevych-2019-sentence}, and BERTopic/embedding-clustering pipelines~\cite{eklund2022topic,grootendorst2022bertopic}; our clustering stage is a local-density heuristic compared against these in Section~\ref{sec:clustering}. A growing body of business work applies ML and LLMs to review analysis, segmentation, satisfaction mapping, and actionable feature extraction~\cite{boughanmi2025reviews,li2025consumer,mustak2024using,praveen2024crafting,teichert2026reviews}, and toward extracting \emph{actionable} feedback insights~\cite{bhattacharjee2022users,mirkovic2025ai,wang2024transit}. For generation we borrow the grounding logic of retrieval-augmented generation~\cite{lewis2021retrievalaugmentedgenerationknowledgeintensivenlp} (without inference-time retrieval, as the representative is fixed by clustering) and use LoRA/QLoRA adaptation~\cite{dettmers2023qloraefficientfinetuningquantized,hu2022lora} over an open Llama~3 model~\cite{grattafiori2024llama}; these are engineering choices, not the contribution. ReviewSense differs from these lines by isolating the evidence-selection front-end for issue-advice generation under a fixed generator, rather than optimizing a topic model or producing a descriptive summary.

\subsection{Prescriptive Analytics and Decision Support}
Prescriptive analytics recommends actions rather than only describing or predicting~\cite{lepenioti2020prescriptive}, and review-specific work optimizes response policies and managerial diagnosis~\cite{yang2019prescribing}. This raises the evaluation bar: absent a user study, advice is judged for specificity, feasibility, and grounding.

\section{The ReviewSense Framework}
\label{sec:method}

\subsection{Overview}
Fig.~\ref{fig:pipeline} summarizes the ReviewSense workflow. Starting from a corpus of customer reviews, we first compute sentence embeddings using the all-MiniLM-L6-v2 sentence-transformer model~\cite{reimers-gurevych-2019-sentence,wang2020minilm}, chosen over larger encoders such as all-mpnet-base-v2 for CPU latency: the pipeline is designed to embed and cluster a $\sim$300-review batch within a few seconds on CPU, without a GPU at inference. We then identify recurrent complaint clusters through iterative cosine-similarity clustering with threshold decay. Evidence is selected either as one representative per cluster for theme-alignment analysis or as a compact set of representatives for external baseline comparison. The selected evidence is passed to a LoRA-adapted Llama 3 8B Instruct model~\cite{grattafiori2024llama,hu2022lora}, which generates structured issue-advice output intended for managerial interpretation.

\subsection{Iterative Semantic Clustering}
\label{sec:clustering}
\begin{figure}[t]
    \centering
    \includegraphics[width=0.92\linewidth]{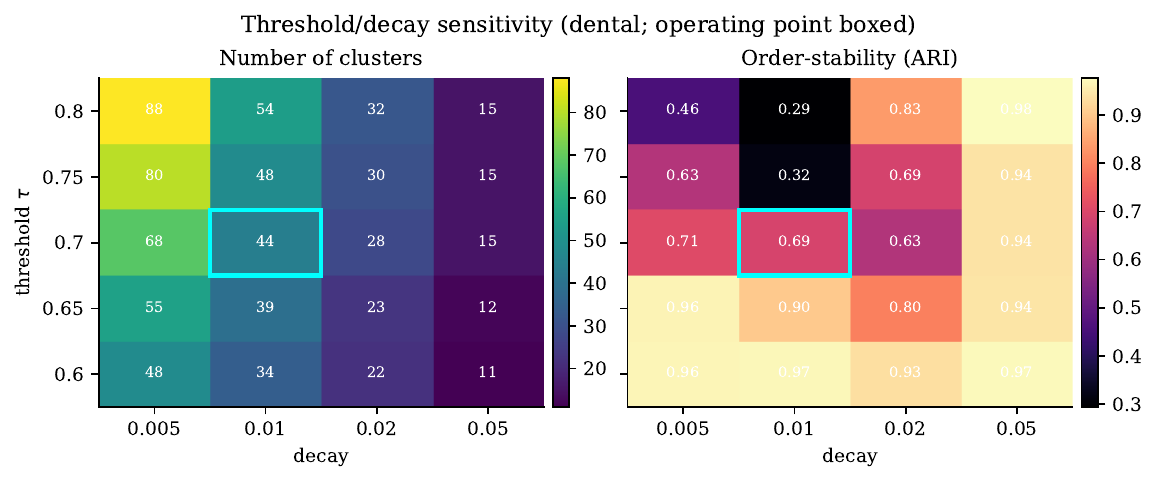}
    \caption{Threshold/decay sensitivity on the dental development split. \emph{Left}: number of extracted clusters (more, finer clusters at higher $\tau$). \emph{Right}: order-stability (mean pairwise ARI over input-order shuffles). Stability is high for $\tau\le0.65$ and declines as $\tau$ rises; the operating point ($\tau{=}0.70$, decay${=}0.01$; boxed) retains moderate stability (ARI${\approx}0.69$) while giving finer, more coherent clusters, and stability collapses for $\tau\ge0.75$.}
    \label{fig:sensitivity}
\end{figure}

Let the input review set be $\mathcal{R}=\{r_1,\dots,r_n\}$ and let $e_i \in \mathbb{R}^d$ denote the sentence embedding of review $r_i$. Pairwise semantic similarity is computed using cosine similarity,
\begin{equation}
s_{ij}=\frac{e_i^\top e_j}{\|e_i\|\,\|e_j\|}.
\end{equation}
For a threshold $\tau$, the neighborhood of review $r_i$ is defined as
\begin{equation}
N_i(\tau)=\{j \neq i \mid s_{ij} \ge \tau\}.
\end{equation}
At each iteration, we select the review whose neighborhood is largest,
\begin{equation}
r^{*}=\arg\max_i |N_i(\tau)|,
\end{equation}
and form a cluster from $r^{*}$ together with all reviews in $N_{r^{*}}(\tau)$. The selected cluster is removed from the remaining corpus, and the similarity threshold is reduced by a small decay factor before the next iteration. In the current configuration, the initial threshold is set to $0.70$ (chosen from candidates $\{0.60,0.70,0.80\}$) and the decay factor to $0.01$, selected by manual inspection of cluster coherence on a development split of the dental domain (Fig.~\ref{fig:sensitivity}); we did not retune these values per test domain, and a principled held-out or cross-domain selection procedure is left to future work.

Algorithm~\ref{alg:cluster} states the procedure precisely. It is deliberately \emph{not} a global clustering objective: there is no loss that is jointly minimized over all assignments. Instead, it greedily extracts the most locally dense complaint group from the remaining reviews, which makes it order-dependent, because the first extracted cluster constrains later ones. We therefore fix three rules that the high-level description leaves implicit. (i) \emph{Stopping}: the procedure halts when no unassigned review has an above-threshold neighbor and the decayed threshold falls below a floor $\tau_{\min}$ (set to $0$ in our experiments, so the threshold decays until no above-threshold neighbour remains; residual reviews form singletons, and generation is conditioned on the top-$K$ densest clusters). (ii) \emph{Tie-breaking}: equal neighborhood sizes are resolved by larger mean intra-neighborhood similarity, then by shorter review length (to counter a bias toward verbose reviews), then by index. (iii) \emph{Ambiguous assignment}: a review within threshold of multiple centers is assigned to the center with the highest cosine similarity, and such multi-membership is logged so that its effect on coverage can be inspected.

\begin{algorithm}[t]
\caption{Iterative threshold-decay complaint extraction}
\label{alg:cluster}
\begin{algorithmic}[1]
\Require reviews $\mathcal{R}$, embeddings $\{e_i\}$, initial threshold $\tau_0$, decay $\delta$, floor $\tau_{\min}$
\Ensure ordered clusters $\mathcal{C}=(C_1,C_2,\dots)$
\State $\mathcal{U} \gets \{1,\dots,n\}$;\ \ $\tau \gets \tau_0$;\ \ $\mathcal{C} \gets ()$
\While{$\mathcal{U} \neq \emptyset$ \textbf{ and } $\tau \ge \tau_{\min}$}
  \For{$i \in \mathcal{U}$}
     \State $N_i \gets \{\, j \in \mathcal{U}\setminus\{i\} : s_{ij} \ge \tau \,\}$
  \EndFor
  \State $r^{*} \gets \arg\max_{i \in \mathcal{U}} |N_i|$ \Comment{ties: mean intra-sim, then shorter $r_i$, then index}
  \If{$|N_{r^{*}}| = 0$}
     \If{$\tau > \tau_{\min}$}
        \State $\tau \gets \max(\tau-\delta,\tau_{\min})$;\ \ \textbf{continue}
     \Else
        \State append each $i\in\mathcal{U}$ as a singleton cluster;\ \ $\mathcal{U}\gets\emptyset$;\ \textbf{break}
     \EndIf
  \EndIf
  \State $C \gets \{r^{*}\} \cup N_{r^{*}}$;\ \ append $C$ to $\mathcal{C}$
  \State $\mathcal{U} \gets \mathcal{U} \setminus C$ \Comment{ambiguous $j$: assign to $\arg\max s_{j,\cdot}$; log multi-membership}
  \State $\tau \gets \tau - \delta$
\EndWhile
\State \Return $\mathcal{C}$
\end{algorithmic}
\end{algorithm}

Removing each extracted cluster before recomputing similarities is deliberate: without it, several clusters tend to form around the \emph{same} dominant complaint because they share most of their reviews, so peeling off a cluster and decaying the threshold forces each subsequent group to surface a distinct, less-dominant issue. This behavior is useful for review analysis because business users often care more about isolating the top recurring problems than about recovering an exhaustive partition of the corpus. It is, however, a heuristic, so standard methods with explicit objectives are natural baselines: $k$-means, agglomerative clustering, HDBSCAN over the same embeddings, LDA~\cite{blei2003latent}, and BERTopic~\cite{grootendorst2022bertopic}. Section~\ref{sec:results} reports a controlled comparison against all of these, together with a stability analysis over input orderings.

\subsection{Representative Review Selection}
\label{sec:repselect}
Each extracted cluster can be reduced to a single representative review or combined into a compact multi-representative context. For the cluster-level theme-alignment analysis, we choose the review with the largest number of above-threshold neighbors in the cluster. Intuitively, this is the most central evidence item under the current similarity rule, and using one representative keeps the LLM input compact and reduces the chance that several partially overlapping complaints are mixed in a single prompt. For the external baseline study, we also evaluate a top-$k$ setting that feeds the representatives of the densest clusters together, testing whether compact multi-evidence conditioning recovers information lost by the single-representative bottleneck.

This choice carries known risks that we make explicit. A \emph{generic} review can share moderate similarity with many others and thus acquire a large neighborhood despite containing little actionable content; \emph{longer} reviews may accrue larger neighborhoods simply because they carry more semantic signal; and the most central review in a mixed cluster may itself be \emph{thematically mixed}, which is one source of the weaker secondary-theme results in Section~\ref{sec:results}. The tie-breaking rules in Algorithm~\ref{alg:cluster} mitigate the length bias only partially. We therefore ablate selection strategies at the cluster level (Section~\ref{sec:results}): a centroid- or medoid-nearest representative is consistently more central and covers its cluster better than the largest-neighborhood choice, whereas the longest-review heuristic inflates both length and genericness. The largest-neighborhood rule is thus an inspectable default for the single-representative condition, while the top-$k$ result in Section~\ref{sec:baselines} suggests that the stronger ReviewSense configuration should use several complementary representatives; downstream medoid/diverse-$k$ evaluation remains future work.

\subsection{Advice Generation}
The generation component is based on Llama 3 8B Instruct~\cite{grattafiori2024llama}, adapted with LoRA~\cite{hu2022lora} for action-oriented output, and is best understood as evidence- conditioned generation~\cite{lewis2021retrievalaugmentedgenerationknowledgeintensivenlp}: the representative review (and its cluster) is the evidence on which the advice is grounded. The model is trained on review-to-advice examples so that it maps free-text complaints to concise issue-advice pairs. For each representative review, the prompt asks the model to (i) identify the core customer issue and (ii) produce a concrete business response, constrained to a short structured form rather than a free-form summary, which keeps outputs easy to inspect. Clustering and generation thus play different roles: clustering isolates recurring evidence from noisy review streams, while the LLM renders that evidence into language a human decision-maker can act on, supporting rather than replacing managerial judgment.

\section{Data and Evaluation}
\label{sec:eval}

\begin{table}[t]
\centering\footnotesize
\caption{Dataset usage. Raw review corpora are proprietary; experiments use negative/neutral 300-review batches after preprocessing.}
\label{tab:data_usage}
\setlength{\tabcolsep}{4pt}\renewcommand{\arraystretch}{1.08}
\begin{tabular}{@{}lrrrr>{\raggedright\arraybackslash}p{3.1cm}@{}}
\toprule
\textbf{Domain} & \textbf{Source} & \textbf{Batch} & \textbf{Clusters} & \textbf{Advice eval.} & \textbf{Role} \\
 & \textbf{reviews} & \textbf{size} & & & \\
\midrule
Dental & 5{,}000 & 300 & 44 & Yes & theme alignment; baselines; clustering \\
Healthcare & 5{,}000 & 300 & 45 & Yes & theme alignment; clustering \\
Real estate & 5{,}000 & 300 & 39 & No & cross-domain clustering \\
\bottomrule
\end{tabular}
\end{table}

\subsection{Review Datasets}
We evaluate ReviewSense on two domain-specific review sets drawn from service businesses in the dental and healthcare sectors (Table~\ref{tab:data_usage}). The source material consists of publicly posted customer reviews collected through an internal analytics workflow. Because the underlying enterprise data are proprietary, we report the domains, processing procedure, and evaluation criteria rather than releasing raw corpora, a reproducibility limitation discussed in Section~\ref{sec:limitations}. In operational use the pipeline processes batches of approximately 300 reviews, balancing issue coverage against runtime. Experimental batches are built deterministically: keep reviews rated 1--3 (negative/neutral), discard empty and near-empty texts, and take the first 300 survivors in corpus order. No deduplication, stratification, or temporal filter is applied. The exported corpora carry no timestamps, so a batch is not a most-recent window. Each domain contributes one batch, leaving batch-to-batch and temporal stability unmeasured. For the clustering analyses in Section~\ref{sec:results} we additionally include a third domain, real-estate (property-rental) reviews, to probe cross-domain generalization beyond the two health-service domains; external advice-generation baselines are currently run on the dental domain.

\subsection{Preprocessing}
Before clustering and generation, the input reviews undergo standard quality-control preprocessing: spam filtering, redaction of personally identifiable information, and basic text normalization. The advice stage prioritizes neutral and negative reviews because these are more likely to expose actionable service failures, unmet expectations, and operational bottlenecks.

\subsection{Evaluation Protocol}
\label{sec:protocol}
Given the decision-support objective, evaluation focuses on thematic alignment and actionability rather than standard summarization metrics. We use two complementary criteria. First, annotators assess \emph{thematic alignment}: for each review set they assign primary and secondary themes to the underlying complaints and then check whether the generated issue-advice outputs preserve those themes. Second, expert reviewers assess \emph{actionability} against a rubric whose dimensions are: specificity (a concrete action rather than a restatement), feasibility (plausible within normal operations), evidence grounding (traceable to the cluster's complaint), operational ownership (who would act), and non-genericness (not applicable to any business). Outputs that were vague, generic, or insufficiently tied to the cluster evidence were treated as weak even when linguistically well formed. Two reviewers with domain familiarity scored the outputs, and theme alignment was annotated internally as an operational quality-assurance pass, not a designed measurement study. The retained export is set-level correct/total counts, no blinding protocol was recorded, and no per-item scores carrying rater identity survive, so inter-rater agreement, paired tests, and confidence intervals cannot be computed retrospectively. We therefore report point estimates only and treat this as the principal evaluation limitation (Section~\ref{sec:limitations}).

\subsection{Results and Discussion}
\label{sec:results}
Table~\ref{tab:theme_alignment_accuracy} reports theme-alignment accuracy across the two domains. The strongest result is that dominant cluster themes are preserved reasonably well: primary-theme alignment reaches 77.3\% when generation is conditioned on one representative review and 79.5\% with two. Secondary-theme alignment is markedly lower, at 60.6\% and 39.4\%, respectively. The domain-level theme-alignment accuracy is computed as
\begin{equation}
\mathrm{Acc}(d,m) =
\frac{\sum_{i \in \mathcal{S}_d} c_{i,m}}
{\sum_{i \in \mathcal{S}_d} n_{i,m}}
\times 100,
\end{equation}
where $\mathcal{S}_d$ is the set of reviewed batches in domain $d\in\{\text{Healthcare},\text{Dental},\text{Overall}\}$ and $c_{i,m}, n_{i,m}$ are the number of correct theme matches and the total number of evaluated comments for batch $i$ under one of the four metrics $m$ (primary or secondary theme alignment, each conditioned on one or two representative reviews).

\begin{table}[t]
\centering
\caption{Theme-alignment accuracy by domain and overall. Parentheses give correct/total counts from the cluster-accuracy worksheet.}
\label{tab:theme_alignment_accuracy}
\small
\setlength{\tabcolsep}{5pt}
\begin{tabular}{lcccc}
\toprule
\textbf{Scope} & \textbf{Primary} & \textbf{Primary} & \textbf{Secondary} & \textbf{Secondary} \\
               & \textbf{(1 rep.)} & \textbf{(2 reps.)} & \textbf{(1 rep.)} & \textbf{(2 reps.)} \\
\midrule
Healthcare & 75.9\% (44/58) & 81.0\% (47/58) & 57.5\% (23/40) & 35.0\% (14/40) \\
Dental     & 80.0\% (24/30) & 76.7\% (23/30) & 65.4\% (17/26) & 46.2\% (12/26) \\
Overall    & 77.3\% (68/88) & 79.5\% (70/88) & 60.6\% (40/66) & 39.4\% (26/66) \\
\bottomrule
\end{tabular}
\end{table}

This gap is the most important empirical finding, and we read it as a limitation of representative-based conditioning rather than a success to be advertised. Conditioning generation on one (or two) central reviews captures the \emph{dominant} complaint signal but under-represents \emph{secondary} concerns in clusters that contain several related issues. Notably, adding a second representative \emph{raises} primary alignment (77.3\%$\rightarrow$79.5\%) but \emph{lowers} secondary alignment (60.6\%$\rightarrow$39.4\%): a second central review tends to reinforce the dominant theme rather than surface a distinct secondary one, crowding the prompt. The practical implication is that ReviewSense is reliable for surfacing and acting on dominant recurring issues, and should remain human-in-the-loop for multi-issue clusters. Richer multi-evidence strategies are the natural way to address this and are left to future work: diverse (rather than merely central) top-$k$ selection, cluster-summary prompting, or retrieval over all cluster members.

\begin{table}[t]
\centering
\caption{Clustering comparison, averaged over the three domains ($N{=}300$ each, same SBERT embeddings; partitional methods matched at the threshold-decay cluster count). Cluster quality is measured \emph{externally} against a lexicon-based category benchmark that is independent of the embedding space. Time is the clustering step only (the shared embedding step is excluded); ``Asg.'' is the percentage of reviews placed in a non-outlier cluster.}
\label{tab:comparison_clustering}
\small
\setlength{\tabcolsep}{4.5pt}
\begin{tabular}{lccccc}
\toprule
\textbf{Method} & \textbf{Time (s)} & \textbf{Asg.\,\%} & \textbf{Homog.} & \textbf{Purity} & \textbf{NMI} \\
\midrule
ReviewSense (threshold-decay) & \textbf{0.01} & 100 & 0.37 & 0.58 & 0.22 \\
$k$-means & 1.06 & 100 & \textbf{0.42} & \textbf{0.62} & \textbf{0.25} \\
Agglomerative (cosine) & 0.09 & 100 & 0.29 & 0.56 & 0.25 \\
LDA & 2.50 & 100 & 0.30 & 0.51 & 0.17 \\
BERTopic & 8.46 & 83 & 0.25 & 0.52 & 0.19 \\
HDBSCAN & 0.14 & 20 & 0.03 & 0.36 & 0.04 \\
\bottomrule
\end{tabular}
\end{table}

With clustering fixed, we benchmark the threshold-decay heuristic against standard alternatives over the same SBERT embeddings on all three domains (Table~\ref{tab:comparison_clustering}); sentence-level SBERT was chosen over graph- and token-level encoders in earlier pilots, so the comparison here is between \emph{clustering algorithms}. Because these corpora have no gold topic labels, we evaluate cluster quality externally, against a lexicon-based category benchmark that is independent of the embedding space, using homogeneity, purity, and normalized mutual information (NMI). The benchmark uses five hand-built keyword categories per domain (billing/insurance, scheduling/wait, staff communication, care or treatment quality, facility or property condition); each review takes the category with the most keyword matches, and reviews matching none are excluded, giving coverage of $94.0\%$, $88.3\%$, and $98.7\%$ for dental, healthcare, and real estate. It is deliberately coarse and unvalidated against human annotation: $63$--$92\%$ of reviews match multiple categories, so hard single-labeling penalizes all methods alike, which is why absolute NMI and ARI are low throughout and only \emph{relative} comparisons should be read from Table~\ref{tab:comparison_clustering}. Two findings stand out. First, the greedy heuristic is by far the cheapest method ($\sim$0.01\,s of clustering vs.\ 0.1--8\,s) and, unlike the density-based methods, it assigns every review (Fig.~\ref{fig:frontier}b); HDBSCAN collapses to 2--3 clusters covering only $\sim$20\% of reviews, and BERTopic finds $\sim$12 topics but still leaves $\sim$17\% unassigned. Second, on the external benchmark ReviewSense is the second most coherent method, behind $k$-means but ahead of agglomerative clustering, LDA, BERTopic, and HDBSCAN, while clustering about two orders of magnitude faster than $k$-means. We also measured order-sensitivity: over 20 random input orderings the heuristic attains a mean pairwise Adjusted Rand Index of $0.72$, well above $k$-means across random initializations ($0.29$) but short of order-invariance, as expected for a greedy procedure. We therefore present threshold-decay clustering not as a state-of-the-art clusterer, since $k$-means is more coherent on every metric, but as a simple, fully-covering local-density heuristic that needs no pre-specified cluster count and yields inspectable recurrent-complaint groups well suited to supplying focused evidence to generation; because end-to-end latency is dominated by LLM generation, its clustering speed is a convenience rather than the central argument. The comparison we do not run is $k$-means followed by the same top-$k$ evidence selection, which would separate the contribution of the clustering stage from that of the evidence-selection stage; we flag it as the most informative next experiment. Throughput and cost at production scale are not evaluated here.

Across successive generator variants, outputs became progressively more concise and specific: Mistral-7B~\cite{Jiang2023Mistral7}, its 4-bit QLoRA~\cite{dettmers2023qloraefficientfinetuningquantized} adaptation, FlashAttention-2~\cite{dao2023flashattention} for speed, and finally LoRA-adapted Llama-3 8B~\cite{grattafiori2024llama}. Table~\ref{tab:strong_weak} contrasts \emph{strong} and \emph{weak} outputs: weak advice is generic or under-specified, whereas strong advice assigns ownership, specifies a mechanism, and stays grounded in the cluster's complaint.

\begin{table}[t]
\centering\footnotesize
\caption{Representative \emph{strong} and \emph{weak} generated advice, with the failure mode for weak cases.}
\label{tab:strong_weak}
\setlength{\tabcolsep}{4pt}\renewcommand{\arraystretch}{1.05}
\begin{tabular}{@{}l p{7.9cm} p{2.4cm}@{}}
\toprule
& \textbf{Generated issue-advice} & \textbf{Judgment} \\
\midrule
Strong & \textbf{ISSUE} wrong quote sent. \textbf{ADVICE} assign an owner to verify each quote (treatment + insurance) before sending; follow up within one day. & Specific; owned; grounded \\
Strong & \textbf{ISSUE} no needle-free option for phobic patients. \textbf{ADVICE} offer topical-anesthetic / laser options and train staff to present them at consultation. & Concrete, feasible \\
Weak & \textbf{ADVICE} expand the array of options available for anxious patients. & Generic; no mechanism/owner \\
Weak & \textbf{ADVICE} review how quotes are managed and ensure the team follows correct procedures. & Restates the complaint \\
\bottomrule
\end{tabular}
\end{table}

At the domain level, dental shows stronger single-representative primary alignment while healthcare benefits from a second representative; expert review echoed the quantitative scores, with practical usefulness depending on both cluster coherence and advice specificity.

\subsection{External Baselines}
\label{sec:baselines}
\begin{figure}[t]
    \centering
    \includegraphics[width=0.92\linewidth]{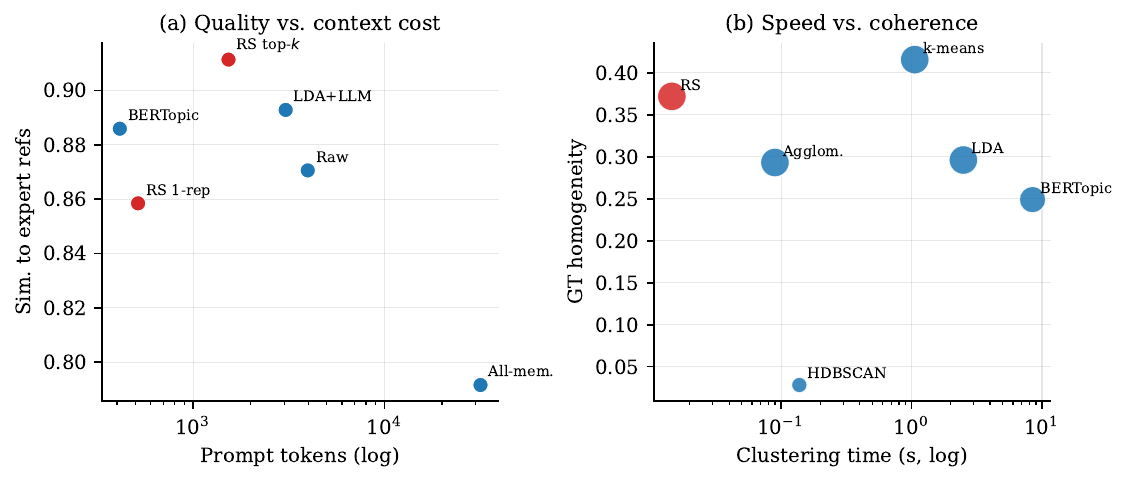}
    \caption{(a) Advice quality versus input-context cost (dental; one fixed generator across all conditions): top-$k$ ReviewSense attains the highest reference-conformity at a fraction of the baselines' tokens, while a single representative is competitive but not the strongest front-end. (b) Clustering runtime versus ground-truth homogeneity over the three domains (marker area $\propto$ \% of reviews assigned): ReviewSense is the fastest method and fully covers the corpus while remaining competitive on coherence, whereas the topic-model baselines are less coherent and HDBSCAN severely under-covers.}
    \label{fig:frontier}
\end{figure}

A central question is whether cluster-guided conditioning actually helps relative to simpler alternatives. We compare ReviewSense against external review-to-advice pipelines that all share the \emph{same} base generator (Llama-3.1-8B, greedy decoding, \emph{no} task adapter, which is distinct from the LoRA-adapted Llama-3-8B of the deployed pipeline, so no fine-tuning advantage accrues to ReviewSense) and the same structured prompt, so that differences reflect only the evidence-selection front-end. Each condition is scored by cosine similarity between its generated advice and an expert recommendation set on the dental domain (Table~\ref{tab:baselines}); this is a \emph{reference-conformity} proxy, not a direct measure of actionability. Two observations follow. First, the strongest configuration is ReviewSense with a small set of top-$k$ representatives ($k{=}5$ in this experiment; 0.911; Fig.~\ref{fig:frontier}), which exceeds every baseline while consuming a fraction of their context. The raw-review and all-members baselines use $3$--$20\times$ more tokens yet score lower, so the benefit comes from \emph{focused} evidence, not more of it. Second, and more soberly, a \emph{single} representative (0.858) is \emph{not} the strongest front-end: it trails LDA+LLM (0.893), BERTopic+LLM (0.886), and even raw-review prompting (0.871). Cluster-guided selection thus helps only when it supplies several complementary representatives; one central review discards too much, consistent with the secondary-theme gap of Section~\ref{sec:results}. We read Table~\ref{tab:baselines} as motivating diverse multi-representative conditioning rather than as a win for single-representative selection. ROUGE-L was uniformly low ($\sim$0.18--0.20) and non-discriminative; a blinded multi-rater actionability study is left to future work. Because the generator here is deliberately non-adapted, Table~\ref{tab:baselines} characterizes evidence-selection \emph{front-ends} and not the deployed end-to-end system, which uses the adapted generator; we do not report the same comparison under that adapted generator.

\begin{table}[t]
\centering
\caption{External baseline comparison for issue-advice generation (dental); one fixed generator across all conditions. ``Tokens'': input-context size; ``Sim.'': maximum cosine similarity of the generated advice to the expert recommendation set (a reference-conformity proxy, not a direct actionability measure); ``R-L'': ROUGE-L against the same references, shown to be uniformly low and non-discriminative.}
\label{tab:baselines}
\small
\setlength{\tabcolsep}{5pt}
\begin{tabular}{llccc}
\toprule
\textbf{Method} & \textbf{Evidence fed} & \textbf{Tokens} & \textbf{Sim.} & \textbf{R-L} \\
\midrule
Direct LLM (raw reviews) & 30 raw reviews & 3{,}974 & 0.871 & 0.18 \\
Direct LLM (all members) & all cluster members & 31{,}767 & 0.792 & 0.20 \\
LDA $+$ LLM & top doc per topic & 3{,}046 & 0.893 & 0.18 \\
BERTopic $+$ LLM & rep.\ doc per topic & 414 & 0.886 & 0.19 \\
ReviewSense (per cluster) & 1 representative / cluster & 516 & 0.858 & 0.19 \\
ReviewSense (top-$k$ reps) & 5 representatives & 1{,}531 & \textbf{0.911} & 0.18 \\
\bottomrule
\end{tabular}
\end{table}

\section{Limitations}
\label{sec:limitations}
Several limitations bound these results. (1) \emph{Evaluation}: theme alignment was annotated as an operational quality-assurance pass, not a designed measurement study. The retained export is set-level, no blinding protocol was recorded, and no per-item scores with rater identity survive, so inter-rater agreement, paired tests, and confidence intervals cannot be computed retrospectively. A blinded study with more annotators and per-item scoring is the most valuable addition to this work. (2) \emph{Scope}: both advice-evaluated domains, dental and healthcare, are health services, so these results are not evidence of generality. Hospitality, restaurant, and e-commerce reviews are untested, real estate contributes clustering evidence only, and each domain contributes one deterministic batch of a proprietary corpus, leaving temporal stability unmeasured. (3) \emph{Evidence baselines}: the top-$k$ gain is not disentangled from evidence quantity and diversity. We do not compare random-$k$ reviews, random or medoid representatives, diversity-aware selection such as maximal marginal relevance, cluster summaries, or raw reviews at an \emph{equal token budget}, nor $k$-means with the same top-$k$ selection, nor recent LLM-based review-analysis systems head-to-head. Because this comparison also fixes a non-adapted generator by design, it characterizes front-ends rather than the deployed system. (4) \emph{Parameters and leakage}: threshold and decay were chosen on a dental development split rather than by cross-domain validation, stability was checked for cluster assignments rather than advice, and no leakage audit was run on a split that is not grouped by business. (5) \emph{Impact and ethics}: we measure reference conformity and rated advice quality, not managerial outcomes, and prioritizing neutral and negative reviews can leave minority complaints unsurfaced while the pipeline does not model review authenticity, fairness across business sizes, or customer consent. These define the agenda: equal-budget evidence baselines, grouped leakage-free splits, blinded multi-rater evaluation, non-health domains, and a decision-impact study.

\section{Conclusion}
\label{sec:conclusion}
ReviewSense is a practical workflow for converting customer reviews into issue-oriented advice for decision support, combining semantic clustering, evidence selection, and LLM generation. Its findings are two-sided: dominant cluster themes are preserved more reliably than secondary ones, and compact multi-representative evidence beats both a single representative and much larger raw contexts, so \emph{which} evidence reaches the generator matters more than how much. Future work should establish equal-budget evidence baselines, blinded multi-rater evaluation, non-health domains, and decision-impact measurement.

\appendix
\paragraph{Reproducibility details.}
The pipeline uses all-MiniLM-L6-v2 embeddings~\cite{reimers-gurevych-2019-sentence,wang2020minilm}; pilots also tried BERT~\cite{devlin-etal-2019-bert}, DistilBERT~\cite{sanh2019distilbert}, RoBERTa~\cite{liu2019roberta}, and all-mpnet-base-v2~\cite{song2020mpnet}. Clustering uses Algorithm~\ref{alg:cluster} with $\tau{=}0.70$, decay $0.01$, and $\tau_{\min}{=}0$. Generation uses Llama-3 8B Instruct~\cite{grattafiori2024llama} with LoRA~\cite{hu2022lora} ($r{=}8$, $\alpha{=}32$, $q/v$, dropout $0.1$), trained for 10 epochs on 1{,}267 review-recommendation pairs with an 80/20 split and 4-bit NF4 inference~\cite{dettmers2023qloraefficientfinetuningquantized}; baselines use base Llama-3.1-8B without adapters and greedy decoding. The prompt requests structured ISSUE/ADVICE output: one issue/advice pair for one representative, or up to five recurring issues for top-$k$ contexts. Raw corpora remain proprietary; the actionability rubric is in Section~\ref{sec:protocol}.

\begin{credits}
\subsubsection{\ackname} This work was supported by Birdeye through the Birdeye AI Program at BITS Pilani. Birdeye provided the proprietary dental, healthcare, and real estate review corpora, as well as the expert recommendation set used for evaluation. These corpora are subject to commercial agreements with the reviewed businesses and therefore cannot be publicly redistributed. Derived aggregate results and evaluation scripts are available from the authors upon request.
\end{credits}

\bibliographystyle{splncs04}
\bibliography{references}

\end{document}